\documentclass{article}
\usepackage{amsmath,amssymb,pifont}
\usepackage{multicol}
\usepackage{amstext}
\usepackage{amsthm}
\usepackage{multirow}
\usepackage{booktabs}
\usepackage[skip=0pt]{subcaption}
\usepackage{times}
\usepackage{lipsum}
\usepackage[shortlabels]{enumitem}
\usepackage{cancel}
\usepackage{wrapfig}
\usepackage{array}
\usepackage{siunitx}
\usepackage{csvsimple}
\usepackage[multidot]{grffile}
\usepackage{bbm}
\usepackage{dblfloatfix}
\usepackage{hyperref}
\usepackage{makecell}
\usepackage{bbm, dsfont}
\usepackage{mathtools}
\usepackage[dvipsnames]{xcolor}
\usepackage{comment}
\usepackage[capitalise]{cleveref}

\usepackage{geometry}
\usepackage{mathpazo}
\usepackage{enumitem}

\usepackage[multiple]{footmisc}
\usepackage{mathrsfs}
\usepackage{todonotes}
\usepackage{tikz}
\usepackage{hyperref}
\usepackage{cleveref}
\usepackage{algpseudocode,algorithm,algorithmicx}
\usepackage{xfrac}

\usepackage{graphicx} 
\usepackage{color}

\usepackage{array}
\usepackage{amssymb}
\usepackage{amsmath}
\usepackage{xspace}
\usepackage{fancyhdr}
\usepackage{comment}
\usepackage{nicefrac}
\usepackage{authblk} 
\usepackage{breakurl}
\usepackage[british]{babel}

\usepackage{fullpage}
\usepackage[numbers, sort, comma, square]{natbib}

\usepackage{xcolor}

\definecolor{darkgreen}{rgb}{0,0.4,0.0}

\PassOptionsToPackage{usenames,dvipsnames}{xcolor}
\usepackage{hyperref} 
\usepackage[capitalise]{cleveref}

\PassOptionsToPackage{sort}{natbib}
\PassOptionsToPackage{square}{natbib}
\PassOptionsToPackage{numbers}{natbib}
\usepackage{natbib}

\usepackage{booktabs}
\usepackage{tabularx}
\usepackage{makecell}
\usepackage{multirow}

\usepackage{ifthen}
\newif\ifieee
\hypersetup{
    colorlinks=true,
    linkcolor=black,
    filecolor=magenta,
    urlcolor=black,
    citecolor=red,
}

\title{Federated Learning in Practice: Reflections and Projections}

 \date{}

\author{
\quad \\
Katharine Daly, \quad Hubert Eichner, \quad Peter Kairouz\thanks{Zheng Xu and Peter Kairouz conceived, coordinated, and edited this work.}, \quad \\ H. Brendan McMahan, \quad Daniel Ramage, \quad Zheng Xu\protect\footnotemark[1] \\
Google Research \\
\{dalyk, huberte, kairouz, mcmahan, dramage, xuzheng\}@google.com
}

\begin{document}
\maketitle
\setboolean{ieee}{false} 

\begin{abstract}
Federated Learning (FL) is a machine learning technique that enables multiple entities to collaboratively learn a shared model without exchanging their local data. Over the past decade, FL systems have achieved substantial progress, scaling to millions of devices across various learning domains while offering meaningful differential privacy (DP) guarantees. Production systems from organizations like Google, Apple, and Meta demonstrate the real-world applicability of FL. However, key challenges remain, including verifying server-side DP guarantees and coordinating training across heterogeneous devices, limiting broader adoption. Additionally, emerging trends such as large (multi-modal) models and blurred lines between training, inference, and personalization challenge traditional FL frameworks. In response, we propose a redefined FL framework that prioritizes privacy principles rather than rigid definitions. We also chart a path forward by leveraging trusted execution environments and open-source ecosystems to address these challenges and facilitate future advancements in FL.
\end{abstract}

\ifthenelse{\boolean{ieee}}{
\begin{IEEEkeywords}
federated learning, differential privacy, cryptography, trusted execution environments
\end{IEEEkeywords}
}{}

\section{Evolution of Federated Learning}
\label{sec:intro} 
Federated Learning (FL) was introduced around 2016 as a privacy enhancing technique that directly applies the principle of data minimization by focused collection and immediate aggregation~\citep{house2012consumer}, which ``\emph{enables mobile phones to collaboratively learn a shared prediction model while keeping all the training data on device, decoupling the ability to do machine learning from the need to store the data in the cloud.}''~\citep{mcmahan2017fedavg,federatedlearn}
FL quickly became a widely-acknowledged paradigm of distributed learning from decentralized data, and has been adopted in various applications beyond the original on-device training scenarios: for example, the FL paradigm has been applied to collaborative learning across multiple institutions (silos) with richer computation resources than mobile devices, or to learning over Internet-of-Things devices with more limited resources. In 2019, out of the discussion in the Workshop on Federated Learning and Analytics at Google, \citet{kairouz2019advances} proposed a broader definition of FL:
\begin{quote}
\emph{\textbf{Federated learning} is a machine learning setting where multiple entities (clients) collaborate in solving a machine learning problem, under the
coordination of a central server or service provider. Each client's raw data is stored locally and not exchanged or transferred; instead, focused updates intended for immediate aggregation are used to achieve the learning objective.}
\end{quote}

Federated Analytics (FA) was introduced later as ``\emph{the practice of applying data science methods to the analysis of raw data that is stored locally on users’ devices. Like FL, it works by running local computations over each device’s data, and only making the aggregated results --- and never any data from a particular device --- available to product engineers.}''~\citep{federatedanalytics} The discussions in this manuscript will primarily focus on FL unless otherwise specified, although they should also be applicable to FA as the two paradigms are closely related to each other and share similar privacy principles~\citep{wang2021fieldguide,bonawitz2021federated}. 

Both FL and FA have made remarkable progress in theory and practice in recent years~\citep{yang2019federated,kairouz2019advances,li2020federated,wang2021fieldguide,elkordy2023federated, sun2024private}. However, despite this progress, production FL systems continue to face a number of existing and new challenges:
\begin{enumerate}
  \item Very large and often multi-modal models achieve unprecedented performance for various tasks, but typically are orders of magnitude beyond what has been considered in classical cross-device FL applications.

  \item Current federated learning systems provide little in the way of verifiability of server-side computations, and even verification of client-side work can be difficult. This limits the ability of a user or external auditor to confirm the privacy properties of the system, and generally the extent to which trust in the service provider can be minimized.

  \item Finally, our practical experience with cross-device FL is that it is often possible but seldom easy. Coordinating training across loosely synchronized and heterogeneous devices (heterogeneous in compute, bandwidth, availability for training, data, and even software version) produces countless operational challenges that hinder the broader adoption of FL.
\end{enumerate}

\begin{table*}
\begin{centering}
\renewcommand{\arraystretch}{1.5}
\caption{A summary of how different privacy principles are addressed under the FL 2017-2020 practice, the FL 2021-2024 practice, and an updated 2025+ goal-state based on the new FL definition we propose. 
}
\label{tab:fl_privacy}
\begin{small}
\begin{tabularx}{\textwidth}{lXXX}
\toprule
       \textbf{\makecell{Privacy \\ Principles}}
       & \textbf{FL 2017-2020}
       & \textbf{FL 2021-2024}
       & \textbf{FL 2025-?} \\
\midrule 
\makecell[t]{Data \\ minimization} & 
Data remain on devices; focused updates and immediate aggregation for model training. &
Trusted and cryptographic aggregation methods can additionally guarantee unaggregated updates invisible to the service provider. &
Secured data on device or cloud with access verifiably limited to specific workloads and immediately revocable (or within a short TTL).
\\
\midrule 
\makecell[t]{Data \\ anonymization} &
No formal anonymization, but messages are collected for the purpose of immediate aggregation. &
Distributed DP can provide acceptable utility for some tasks, and protection from an honest-but-curious service provider;  central DP can provide better utility, and strong DP protection for the model released to end users but assumes a trusted aggregator. &
Achieve the utility of current Central DP approaches, while also offering strong protection against even a malicious service provider;
users can verify that only anonymized results are released, and can enforce their privacy preferences.
\\
\midrule 
\makecell[t]{Transparency \\ and \\ control} &
Users can choose whether to participate in training, and potentially inspect the on-device binaries and network usage. &
Users can additionally inspect the source code of \textit{some} FL instances such as Private Compute Core~\citep{marchiori2022android}, while others remain closed source and proprietary. 
&
Users can view a human-readable summary of the purpose and (privacy) properties of any computation their data participated in, and those properties can be verified. Users can make fine-grained choices about which FL workloads to run, or delegate that power to  an organization of their choice. 
\\
\midrule 
\makecell[t]{Verifiability \\ and \\ auditability} &
Where code is open-sourced, it can be inspected; verifying the identity of the code running on devices is possible but difficult. &
Same as FL 2017-2020 &
Client and server-side code verify each others’ integrity via remote attestation. Clients can verify the data minimization and anonymization properties of server-side computation. Clients and servers verify each others' authenticity via (ideally independent) Public Key Infrastructure (PKI).  \\
\bottomrule
\end{tabularx}
\end{small}
\end{centering}
\end{table*}
To facilitate the advancement of next generation federated technologies considering the above-mentioned challenges and opportunities, we revisit the defining characteristics and propose a new definition of FL, which aims not to draw a hard line between what ``is" and ``isn't" FL, but rather highlight the principles and aspirations of research and infrastructure. Before proceeding, we first review the privacy principles initially presented by \citet{bonawitz2021federated}: 
\begin{enumerate}
    \item The user has \textit{transparency, auditability, and control} over what data is used, what purpose it is used for, and how it is processed. This includes forward-looking transparency, retrospective auditability of computation or release details, control of at least the immediate use of data (e.g. in training) in addition to others. 
    \item Processing of user data (whether training examples or gradients) should encode \textit{data minimization} by reducing the information any actor has access to at every node in the system. This includes things like sending only focused, minimal updates back to the service provider (rather than raw data), aggregating the updates in memory, sharing only select updates with the engineers that have requested the computation, and using secure enclaves and/or cryptographic primitives to hide potentially sensitive data from various actors in the system. 
    \item Released outputs should provide formal 
    \textit{data anonymization} guarantees, ensuring that released outputs do not reveal anything unique to an individual. In other words, aggregate statistics, including model parameters, when released to an engineer (or beyond) should not vary significantly based on whether any particular user's data was included in the aggregation. 
    \item Privacy claims are \textit{verifiable} ideally by the users themselves, by external auditors, and the service provider. 
\end{enumerate}
To more effectively capture the aforementioned privacy principles and address the outlined challenges, we propose the following new definition:
\begin{quote}
\emph{\textbf{Federated learning (FL)} is a machine learning setting where multiple entities (clients) collaborate in solving a machine learning problem, under the coordination of a service provider. A complete FL system should enable clients to maintain full control over their data, the set of workloads allowed to access their data, and the anonymization properties of those workloads. FL systems should provide appropriate transparency and control to the users whose data is managed by FL clients.}
\end{quote}
One goal of this new definition is to focus more on the privacy properties of the system, rather than how they are obtained. For example, ``appropriate transparency and control'' could be maintained while allowing users to delegate workload or privacy choices to a trusted third party other than the service provider.
Even with this definition, claiming a particular system is ``doing FL" is not (and has never been) sufficient to provide a full picture of its specific privacy properties; rather a more nuanced and detailed statement is necessary, highlighting how the system approaches the multi-faceted privacy principles mentioned above. \cref{tab:fl_privacy} describes how different facets of privacy are addressed by the traditional (2019) FL definition, typical cross-device FL practice, and in an ideal north-star version of FL.

The remainder of the paper is organized as follows. Section \ref{sec:practice} discusses the major advances in practice with a heavy focus on Google's FL technology. Section \ref{sec:challenges} presents the remaining challenges and emerging opportunities. Section \ref{sec:future} charts a path forward by proposing a new design paradigm for federated learning. We conclude the paper in Section \ref{sec:conclusion}.

\section{Advances in Federated Learning in Practice}\label{sec:practice}

In recent years, practical FL systems have harnessed significant advancements by the community: we can scale to millions of devices and many domains; we can apply secure multiparty computation protocols at scale and combine them with central or distributed DP; we can successfully train production models with meaningful DP guarantees while achieving high utility. In this section, we summarize recent progress of FL in practice by reexamining the open problems in FL, taking a retrospective view inspired by \citet{kairouz2019advances}. Rather than conducting an exhaustive review of recent publications, we emphasize the practical developments and highlight avenues where more research is needed. The discussions heavily focus on the progress in industry applications built on large-scale systems that are primarily consolidated from the keynote talks and discussions from the Federated Learning and Analytics in Practice Workshop~\citep{xu2023federated}, and biased to cross-device federated learning (compared to cross-silo or other settings) due to the familiarity of the authors. 

\paragraph{Applications} At Google, FL has been applied to training several machine learning models powering advanced features in mobile keyboard (Gboard) including next word prediction \citep{hard2018gboard,xu2023gboard, meta_fl}, smart compose and on-the-fly rescoring for suggestions \citep{xu2023gboard}, and emoji suggestion \citep{gboard19emoji}.  Some additional applications include keyword spotting model for virtual assistants \citep{hard2022production}, smart text selection on Android \citep{ss_blogpost, hartmann2023distributed},  smart reply and other assistive suggestions in Android Messages \citep{androidmessages2020}, and improving user experience on Pixel phones \citep{googledps2020}. FA has been applied to Google Health Studies to power privacy-preserving health research~\citep{ghs2020}, and in Apple Photos to identify iconic scenes \citep{apple2023}.

\paragraph{Systems} 
Several production systems have been built and discussed, e.g., the cross-device federated system at Google \citep{bonawitz2019towards}, Apple \citep{paulik2021federated, mcmillan2022private, talwar2023samplable}, Meta \citep{huba2022papaya,stojkovic2022applied}. 
Large-scale cross-device systems such as these share challenges related to computation and resource constraints, including: limited server-side control over client participation because devices can only train when they meet (restrictive) local criteria (e.g. being connected to an unmetered network and having appropriate power/charging and idle status); limited and heterogeneous computation power per device; and limited bandwidth as well as relatively high likelihood of dropping out mid-computation. Real-world systems have developed different approaches to tackle the client scheduling challenges from intermittent connection and stragglers. For example, \citet{bonawitz2019towards} used oversampling and dropout, and \citet{huba2022papaya} used asynchronicity.

\paragraph{Privacy}
Federated learning realizes the data minimization privacy principle~\citep{bonawitz2021federated} in collaborative learning, and can combine with other techniques to strengthen privacy protection. For example, secure aggregation methods are used to enhance guarantees for data minimization, and differential privacy (DP) methods are used to provide data anonymization guarantees. Single-server secure aggregation (SecAgg) \citep{bonawitz2017practical} can guarantee that an honest-but-curious server can only observe the aggregated updates derived from many users instead of viewing each individual update. An efficient SecAgg algorithm \citep{bell2020secure} has been developed to scale up to aggregating updates from models of millions of parameters and thousands of clients per round, which is applied in practice to train Gboard language models and Android smart selection models \citep{xu2023gboard,zhang2023private,hartmann2023distributed}. Distributed DP, where the clients can locally add noise and the honest-but-curious server will aggregate the noisy updates, has been applied to train the smart selection models \citep{hartmann2023distributed}. However, the data anonymization is only examined by empirical privacy auditing of the Secret Sharer methods~\citep{carlini2019secret} as the noise added is too small to provide meaningful formal DP guarantees in the cross-device federated systems that cannot perform random sampling for privacy amplification. DP-FTRL \citep{kairouz21practical,choquette2023amplified,mcmahan2024hassle} with stateful noise mechanisms on the server can be used to achieve meaningful formal guarantees when assuming the server will honestly add noise to the aggregated updates, and is applied to train and launch more than thirty Gboard language models with $(\epsilon, \delta)$-DP of $\epsilon \in [0.994, 13.69]$ and $\delta=10^{-10}$ (alternatively, $\rho-$zCDP~\citep{bun2016concentrated} of $\rho \in [0.0144, 1.86]$)~\citep{dpftrl_blogpost,xu2023gboard}.

\paragraph{Algorithms} FL highlights the need of system and algorithm co-design. The federated averaging (FedAvg) algorithm~\citep{mcmahan2017fedavg} and its variants~\citep{wang2021fieldguide} are among the most popular algorithms in practice. The generalized FedAvg variants consider the two stage optimization framework: clients perform local updates on private data with client optimizers, and the server will apply the aggregated update from multiple clients with a server optimizer. In addition to the communication efficiency benefits in the real-world federated system, the FedAvg framework makes it easy to take advantage of the progress in centralized training, and combine with other privacy techniques. For example, adaptive optimizers can be used on the server \citep{reddi2021adaptive} to significantly improve performance of language tasks; adaptive optimizers can also be used on the clients when resources support slightly heavier computation \citep{wang2021local}; when combining with local operators like clipping, DP-SGD~\citep{abadi2016deep,mcmahan18learning} or DP-FTRL~\citep{kairouz21practical,mcmahan2024hassle} can be used as server optimizer to achieve differential privacy.

\section{Challenges and Opportunities} \label{sec:challenges}

The previous section outlined significant advancements in deploying federated learning systems across various domains. However, despite this progress, several key challenges remain, which we will examine in detail in this section.

\subsection{Scaling to Large Foundation Models}
 Recently, large foundation models \citep{bommasani2021opportunities} have attracted much attention in both academia and industry, and have drastically changed the machine learning paradigm. Such models (e.g., OpenAI GPTs~\citep{OpenAI_GPT4_2023} and Dall-Es~\citep{shi2020improving}, Google PaLMs~\citep{anil2023palm} and Geminis~\citep{team2023gemini}, and Meta Llamas~\citep{touvron2023llama,dubey2024llama}) have very large parameter size that can easily scale up to hundreds of billions parameters, and are pre-trained with a very large dataset that can have (tens of) trillions of tokens. The scale of the model and data are both much larger than previous deep learning applications and what has been explored in cross-device federated learning. 

 Such foundation models are strong few-shot and zero-shot learners and can accomplish various tasks with the help of instruction tuning and prompt engineering~\citep{brown2020language,wei2022chain,ouyang2022training,wei2021finetuned}, outperforming previous domain specific smaller models. The development of large language models relies heavily on extensive, high-quality user data, underscoring the growing importance of privacy-preserving techniques in the training process. Here are four primary avenues for incorporating user data into foundation models. 
 \begin{enumerate}
     \item Post-training, popularized by instruction tuning~\citep{ouyang2022training} that combines supervised fine-tuning and RLHF techniques, has become a standard for training large models. User instructions are crucial for aligning large models, but may also contain sensitive private information~\citep{yu2024privacy} that can be memorized~\citep{nasr2023scalable}.
     \item User data can be particularly helpful when adapting foundation models to specific domains, for example, for medical usage \citep{singhal2023towards,bosselut2024meditron}. For improving user typing experience in virtual keyboard, early experiments~\citep{wu2024prompt} suggest the current practices of leveraging large models still cannot compete with what can be achieved by privacy-preserving training with user data. Fine-tuning pre-trained large foundation models on domain specific user data is therefore important but has been shown to carry several privacy risks \cite{kandpal2023user}.
     \item There is a growing interest in training smaller foundation models of billions parameters instead of tens of billions parameters to reduce serving cost and inference latency, and deploying on-device to improve privacy. Early experiments~\citep{cho2024heterogeneous,yu2024privacy} suggest high-quality in-domain data can be used to close the gap between large and small foundation models. 
    \item In addition, there are concerns that foundation models have exhausted the available public data on the web, and the public data will be more and more polluted by hallucinated content generated by current large models~\citep{sani2024future}. 
 \end{enumerate}

Federated learning of large language models is an active research topic, with several surveys released in the last two years \citep{yao2024federated,chen2023federated,woisetschlager2024survey,zhuang2023foundation,yu2023federated}. While researchers have been working hard to develop new algorithms scaling up the model size in FL, the current FL system can only reliably train models with millions of parameters in practical applications (especially in the cross-device FL setting, see \cref{sec:practice}). 
We highlight challenges in scaling to large foundation models in FL.
    The communication and computation resource requirements have been important considerations through the multi-year development of FL. More recently, we have observed for cross-device FL that computation and memory constraints of mobile devices have become the main bottleneck for training large models. Large foundation models bring this challenge to the next level.  
The opportunities of private training for LLMs and challenges of on-device training motivate us to rethink the design of federated learning systems.

\subsection{Verifying Server-side Privacy Guarantees}
As discussed in the introduction, protecting the privacy of users that participate in federated training is of utmost importance since FL's primary motivation is privacy. We now turn our attention to describing the remaining challenges in this space. 

The first generation of FL algorithms and systems (referred to as FL 2017-2020) offered data minimization but still suffered from the possibility of exposing private information through model updates, which can be exploited by a malicious service provider. Indeed, without proper safeguards, a dishonest or compromised service provider could analyze unaggregated updates to infer private details about individual participants \citep{boenisch2023curious, suliman2023two}. 

Since then, several techniques have been developed to mitigate some of these risks, including secure multiparty computation (SMPC) schemes, such as those based on honest-majority cohorts \citep{bonawitz2017practical}, non-colluding secure aggregators \citep{talwar2023samplable}, and hardware-based trusted execution environments (TEEs) \citep{huba2021papaya}. These methods strengthened the data minimization guarantees and ensured that an honest-but-curious server\footnote{An honest-but-curious server is one that follows the protocol but could try to gain insights about users from data it receives. This models the situation where an attacker can't alter an execution on the server, e.g. write and deploy new code to implement an attack, but might store and post-process all data received by the server.} can only see aggregated model updates.  

Another important development that happened is incorporating data anonymization in federated system by using differential privacy (DP) \citep{dwork2006calibrating}. Recent work \citep{xu2024dpmfblog, xu2023federated} has demonstrated the feasibility of training high-utility models with DP, ensuring that model parameters remain statistically indistinguishable whether or not a particular device’s data is included. 

We have also seen attempts at combining data minimization and data anonymization techniques. For example, distributed DP based FL systems \citep{kairouz2021distributed, agarwal2021skellam, hartmann2023distributed} combine single-server secure aggregation protocols with on-device noise to ensure that the service provider can only see a differentially private aggregate. Under distributed DP, clients first compute minimal application-specific reports, perturb these slightly with random noise, and then execute a private aggregation protocol. The server then has access only to the output of the private aggregation protocol. The noise added by individual clients is typically insufficient for a meaningful local DP guarantee on its own. After private aggregation, however, the output of the private aggregation protocol provides a stronger DP guarantee based on the total sum of noise added across all clients. This applies even to someone with access to the server under the security assumptions necessary for the private aggregation protocol. 

Distributed DP represented a major leap forward in improving the privacy guarantees of an FL system. However, distributed DP algorithms suffer from principal limitations that stem from the complexities involved in implementing state-of-the-art DP mechanisms in a distributed setting. These mechanisms either require complex random device sampling protocols, which are difficult to achieve securely in a distributed environment \citep{talwar2023samplable}, or depend on statefulness \citep{mcmahan2024hassle, kairouz21practical}, which poses additional implementation challenges. Consequently, a notable performance gap remains between centralized DP models and distributed DP models.

Another critical challenge in achieving robust verifiable privacy guarantees lies in ensuring resilience against Sybil attacks \citep{douceur2002sybil}. In such attacks, a malicious service provider could inject specially crafted messages into the secure aggregation process to extract sensitive information about a specific individual. Developing scalable and robust defenses against this sort of vulnerability, particularly in SMPC-based secure aggregation schemes, remains an open problem.

While substantial progress has been made in training models with meaningful differential privacy in federated settings, further work is needed to ensure external verification of these privacy guarantees. Addressing the gap between centralized and distributed DP, as well as mitigating the risks posed by adversarial behaviors, will be crucial for the continued adoption and trustworthiness of federated learning in production systems.

\subsection{Addressing System Challenges}
The last few years saw several large-scale deployments of federated systems from various companies, including Google \citep{bonawitz2019towards}, Apple \citep{paulik2021federated, mcmillan2022private, talwar2023samplable}, and Meta \citep{huba2022papaya,stojkovic2022applied}. 
Google's cross-device federated learning system \citep{bonawitz2019towards} features various synchronization points - in part to support the synchronous, round-based FedAvg learning algorithm~\citep{mcmahan2017fedavg}, in part to aid data minimization by supporting the secure aggregation protocol \citep{bonawitz2017practical}. Notably, \citep{huba2022papaya} propose an asynchronous system instead, in large part due to system design considerations; here we elaborate on the challenges faced by the synchronous system introduced in \citep{bonawitz2019towards}.
Specifically, \textit{cohort formation} - collecting a set of devices that execute a federated computation - and aggregation represent points where the system blocks and a decision has to be made when to proceed or fail.
Blocking means in most cases keeping devices waiting, which is inefficient, increases the probability of devices dropping out and hence downstream failures, and can induce bias \citep{kairouz2019advances}. Hard cut-offs - or making the associated proceed/fail decision - leads to a variety of problems in understanding and therefore debugging, maintaining or improving the system:
\begin{enumerate}
    \item Hard cut-offs lead to bifurcation points (phase transitions). Like in non-linear dynamical systems or deep neural nets, small upstream changes can induce sudden / large qualitative downstream changes; likewise, large upstream changes may not have any expected downstream effects.
    \item Synchronization points are all potential failure points - places where computations can fail under \textit{normal operation} because some timeout is hit or a threshold is not reached, significantly complicating debugging because an entire class of errors may be fine and not indicate a real problem.
    \item Lots of knobs make the system harder to operate, monitor, and optimize: time-outs, or thresholds (e.g., reporting goal for minimum number of participating clients every round), and overallocation of devices to mitigate dropout lead to more telemetry, documentation, and require system understanding.
    \item Synchronization points imply coordination across components, leading to complex architectures, cascading errors and network effects.
    \item Synchronization points can cause problems for A/B experiments; while a typical set up would split e.g. devices into control and treatment groups that differ in one setting and are independent of each other, synchronization points introduce dependencies and thus violate the assumption behind A/B experiments.
\end{enumerate}

\section{A Path to the Future} \label{sec:future}
Building upon years of development for advanced FL (\cref{sec:practice}), we discuss a potential path towards the future that instantiates the new FL definition (\cref{sec:intro}) to address challenges in \cref{sec:challenges}. There is growing interest in confidential cloud computation based on hardware and encryption for privacy and security, including for private inference \citep{apple2024pcc} and federated learning systems \cite{huba2022papaya}.

\begin{figure*}[ht!]
    \centering
    \includegraphics[width=0.8\textwidth]{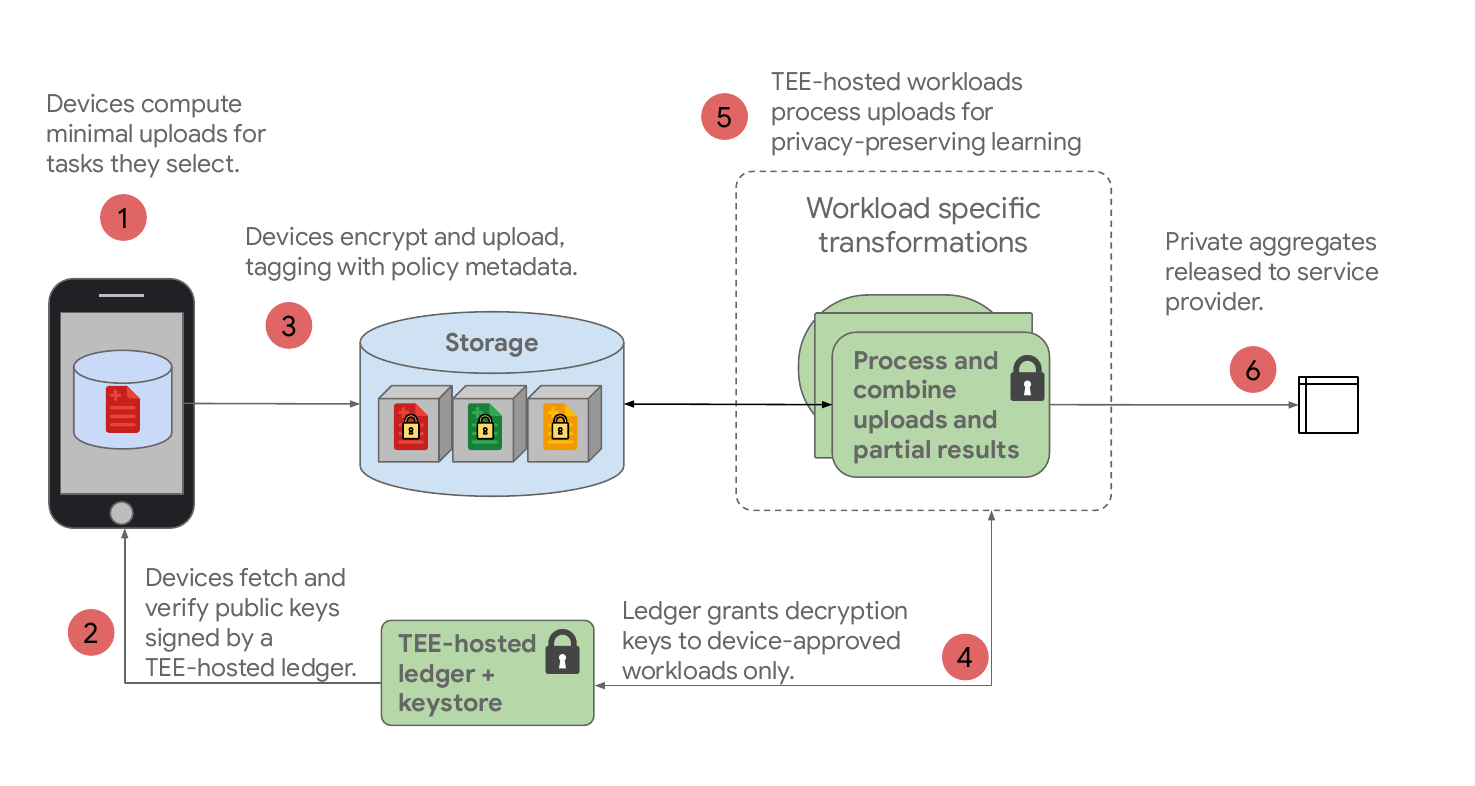}
    \caption{A prototype architecture for using confidential federated computations to train large models. In contrast to traditional cross-device federated learning, devices upload (re-processed and encrypted data, and an iterative training process is performed on the server. An orchestrator is responsible for passing encrypted blobs amongst storage locations and components running in TEEs. The ledger enforces that workload specific transformations adhere to the access policy associated with data uploads. 
    }
    \label{fig:fl_diagram}
\end{figure*}

Our recent work in \citet{eichner2024confidential} proposed a system design for confidential federated computation that leverages Trusted Execution Environments (TEEs) to significantly improve privacy claims with external verifiability, while simultaneously improving system robustness and scalability. In contrast to earlier designs, confidential federated computations allow the device to verifiably limit any server-side processing of uploaded messages to a fixed, known set of approved, privacy preserving workloads. Before upload, devices encrypt messages with a public key whose private key is held by a TEE-hosted ledger service. Devices verify that the public keys are generated by a ledger binary built from known OSS source code and running on a physical TEE with known confidentiality and integrity guarantees. The ledger, in turn, enforces that decryption keys are given only to workflow binaries consistent with a device-approved access policy associated with the message at upload time. The ledger does so by confirming that the workflow binaries are built from approved OSS source code and running on a physical TEE, following the same procedure as the device used to verify the ledger's integrity.

Confidential federated computations can thereby establish an externally verifiable chain of trust, where messages uploaded to the server can be decrypted only in accordance with a device-approved access policy consisting of a graph of permitted transformations on the uploaded data. These properties can be checked by anyone with access to the source code, which can include the general public when devices enforce that the ledger and workloads are reproducibly built from OSS components. Devices retain complete control over what data processing steps can be applied to uploaded data, including requiring specific data minimization or anonymization constraints prior to release of data derived from device uploads. For example, the device might require that a federated learning workflow combine intermediate aggregates with a differentially private algorithm like \cite{mcmahan2024hassle}, releasing only DP model parameters to the service provider.

In this way, confidential federated computations promise solutions to some privacy and system challenges discussed in \cref{sec:challenges}. The application to cross-device federated learning described in \citet{eichner2024confidential} allows cross-device confidential federated computations to remove synchronicity requirements at upload and aggregation time, allowing for better scaling across clients while preserving federated learning's standard approach to data minimization \citep{bonawitz2021federated}. However, even confidential cross-device federated computations require clients to be online to contribute to model training, retaining the well-known bias issues introduced by client heterogeneity. And it is still limited to models of $\sim$10-20M parameters, the current practical bound determined by network and device constraints. Applying methods like LoRA \citep{hu2021lora} or prompt tuning \citep{lester2021power} to FL \citep{cho2023heterogeneous,collins2023profit} could increase the trainable model size, but training models up to billions of parameters remains out of reach even when adopting parameter efficient training.

Unlike traditional cross-device federated learning, confidential federated computations have the potential to apply to large language models and other generative artificial intelligence models. By utilizing the chain of trust, resource-intensive and round-dependent per-client computation such as gradient computation can be moved from mobile devices to TEEs as shown in Figure \ref{fig:fl_diagram}, while preserving externally verifiable differential privacy. For example, the device can verify that the server applies differentially private aggregation properly after per-client model updates in TEEs, and all workload specific transformations adhere to an access policy. In both the traditional and updated notion of FL, per-device information is never visible to the service provider, now enforced via encryption and TEEs rather than via on-device computation placement. Access policies could also be used to provide external verifiability of other kinds of private training algorithms, including techniques that use batches of data spanning multiple clients, which facilitates the integration of the latest centralized training practice. By uploading pre-processed and encrypted data, communication costs are reduced compared to previous FL practice.

Confidential federated computations offer the ability to train much larger models in more flexible ways but there are significant challenges. For one, access policies should be able to enforce correct application of stateful DP algorithms \citep{choquette2023amplified,mcmahan2024hassle} in horizontally-scaled deployments (multiple worker TEEs being used in parallel), yet access policies should always remain sufficiently straightforward for humans to interpret such that they are convincing to external researchers wishing to validate the access policy's guarantees. Second, some DP algorithms require that the orchestrator responsible for passing information between TEEs does not know which devices are participating in each round  \citep[Table 3]{ponomareva2023dpfy}). Special care is required to identify such additional constraints and encode them into the access policy and/or OSS TEE binaries such that they can be externally verified alongside the more straightforward logical transformations. Third, TEE integrity bugs or side-channel leakage to the service operator have the potential to expose private data even if the other aspects of the system are working correctly. Finally, there is overhead associated with running logic in TEEs and potential bottlenecks associated with the ledger in our system. These are some of the many challenges we hope to explore with the community as federated learning and analytics evolve to incorporate new server-side hardware capabilities.

\section{Conclusion} \label{sec:conclusion}
Since its introduction, federated learning has evolved significantly in its practical applicability as well as by incorporating complementary privacy technologies such as differential privacy.  But challenges in the field remain. This paper outlines some scalability and system challenges, especially concerning large foundation models, as well as the need for verifiability of server-side privacy guarantees. This work proposes a new definition of FL to address the evolving landscape of technologies and applications, prioritizing privacy principles over computation placement.

We are excited about a path forward for federated learning systems that utilize confidential cloud computation paired with on-device computation to surpass previous system limitations and unlock new possibilities for privacy-preserving collaborative learning. We invite the research community to explore the possibilities for federated algorithms and systems in federated learning's next chapter.

\section*{Acknowledgements}
 In July 2023, the Federated Learning and Analytics in Practice Workshop~\citep{xu2023federated} brought together academics and practitioners to exchange ideas; discuss systems and applications to inspire research that could lead to real-world impact; and to identify promising future directions. We thank the workshop participants for their engagement and post-workshop discussions, which inspired an earlier draft of some of the ideas in this paper. This paper reflects the views of its authors.
 
 We thank Adria Gascon and Kallista Bonawitz for discussing and reviewing an early draft. We thank the Google federated learning team members and Gboard production partners for various discussions that are partially reflected in this paper. 

\bibliographystyle{abbrvnat}
\small
\bibliography{refer}

\end{document}